\documentclass[conference]{IEEEtran}
\IEEEoverridecommandlockouts
\usepackage{amsmath,amssymb,amsfonts}
\usepackage{algorithmic}
\usepackage{graphicx}
\usepackage{textcomp}
\usepackage{xcolor}
\def\BibTeX{{\rm B\kern-.05em{\sc i\kern-.025em b}\kern-.08em
    T\kern-.1667em\lower.7ex\hbox{E}\kern-.125emX}}

\usepackage[
backend=biber,
style=ieee,
sorting=none,
url=false,
doi=false,
isbn=false,
eprint=false
]{biblatex}

\bibliography{ref.bib}

\DeclareFieldFormat{journaltitle}{#1\isdot}
\DeclareFieldFormat{conferencetitle}{#1\isdot}
\DeclareFieldFormat{title}{#1\isdot}

\usepackage{fancyhdr}
\fancypagestyle{plain}{
  \fancyhf{}
  \fancyfoot[C]{\thepage}%
}
\fancypagestyle{firstpage}{
  \fancyhf{}
  \fancyfoot[C]{\small © 2022 IEEE. Personal use of this material is permitted. Permission from IEEE must be obtained for all other uses, in any current or future media, including reprinting/republishing this material for advertising or promotional purposes, creating new collective works, for resale or redistribution to servers or lists, or reuse of any copyrighted component of this work in other works.}%
}

\begin{document}

\title{In Search of Deep Learning Architectures for Load Forecasting: A Comparative Analysis and the Impact of the Covid-19 Pandemic on Model Performance
}

\author{
\IEEEauthorblockN{\hspace{-1.3cm}Sotiris Pelekis}
\IEEEauthorblockA{
\textit{\hspace{-1.3cm}Decision Support Systems Laboratory} \\
\textit{\hspace{-1.3cm}National Technical University of} \\
\textit{\hspace{-1.3cm}Athens, ICCS, Greece} \\
\hspace{-1.3cm}0000-0002-9259-9115}
\and
\IEEEauthorblockN{\hspace{-1cm}Evangelos Karakolis}
\IEEEauthorblockA{
\textit{\hspace{-1cm}Decision Support Systems Laboratory} \\
\textit{\hspace{-1cm}National Technical University of} \\
\textit{\hspace{-1cm}Athens, ICCS, Greece} \\
\hspace{-1cm}0000-0003-2833-3088}
\and
\IEEEauthorblockN{\hspace{-0.3cm}Francisco Silva\hspace{+0.8cm}}
\IEEEauthorblockA{
\textit{\hspace{-0.3cm}R\&D NESTER\hspace{+0.8cm}} \\
\textit{\hspace{-0.3cm}Lisbon, Portugal\hspace{+0.8cm}} \\
\hspace{-0.3cm}francisco.silva@rdnester.com\hspace{+0.8cm}}
\and
\IEEEauthorblockN{\hspace{+0.2cm}Vasileios Schoinas}
\IEEEauthorblockA{
\textit{\hspace{+0.2cm}Decision Support Systems Laboratory} \\
\textit{\hspace{+0.2cm}National Technical University of} \\
\textit{\hspace{+0.2cm}Athens, Greece} \\
\hspace{+0.2cm}0000-0001-9809-6220}
\and
\IEEEauthorblockN{Spiros Mouzakitis}
\IEEEauthorblockA{
\textit{Decision Support Systems Laboratory} \\
\textit{National Technical University of} \\
\textit{Athens, ICCS, Greece} \\
0000-0001-9616-447X}
\and
\IEEEauthorblockN{Georgios Kormpakis}
\IEEEauthorblockA{
\textit{Decision Support Systems Laboratory} \\
\textit{National Technical University of} \\
\textit{Athens, ICCS, Greece} \\
0000-0003-4052-4549}
\and
\IEEEauthorblockN{\hspace{1.1cm}Nuno Amaro}
\IEEEauthorblockA{
\textit{\hspace{1.1cm}R\&D NESTER} \\
\textit{\hspace{1.1cm}Lisbon, Portugal} \\
\hspace{1.1cm}nuno.amaro@rdnester.com }
\and
\IEEEauthorblockN{\hspace{-1.9cm}John Psarras}
\IEEEauthorblockA{
\textit{\hspace{-1.9cm}Decision Support Systems Laboratory} \\
\textit{\hspace{-1.9cm}National Technical University of} \\
\textit{\hspace{-1.9cm}Athens, ICCS, Greece} \\
\hspace{-1.9cm}john@epu.ntua.gr}
}

\maketitle

\begin{abstract}
In power grids, short-term load forecasting (STLF) is crucial as it contributes to the optimization of their reliability, emissions, and costs, while it enables the participation of energy companies in the energy market. STLF is a challenging task, due to the complex demand of active and reactive power from multiple types of electrical loads and their dependence on numerous exogenous variables. Amongst them, special circumstances—such as the COVID-19 pandemic—can often be the reason behind distribution shifts of load series. This work conducts a comparative study of Deep Learning (DL) architectures—namely Neural Basis Expansion Analysis Time Series Forecasting (N-BEATS), Long Short-Term Memory (LSTM), and Temporal Convolutional Networks (TCN)—with respect to forecasting accuracy and training sustainability, meanwhile examining their out-of-distribution generalization capabilities during the COVID-19 pandemic era. A Pattern Sequence Forecasting (PSF) model is used as baseline. The case study focuses on day-ahead forecasts for the Portuguese national 15-minute resolution net load time series. The results can be leveraged by energy companies and network operators (i) to reinforce their forecasting toolkit with state-of-the-art DL models; (ii) to become aware of the serious consequences of crisis events on model performance; (iii) as a high-level model evaluation, deployment, and sustainability guide within a smart grid context.

\end{abstract}


\section{Introduction}
\thispagestyle{firstpage}
\subsection{Background}
Electricity load forecasts are crucial in assuring the development and optimal operation of power systems, which have always been a significant challenge for any country regardless of their economical state. Load forecasts, both at demand and generation side, can serve in maintaining the balance between load and generation, the economic power dispatch, storage scheduling, network planning and expansion of power grids. Recently, this task is becoming even more popular \cite{Hong2019a} and important, especially for European countries as the energy crisis has reached an unprecedented peak with geopolitical extensions. According to the European Union’s Institute for Security Studies \cite{EuropeanUnionInstituteforSecurityStudies2022EuropesConundrum} the balance between the security, affordability and sustainability dimensions of the energy "trilemma" has come under serious strain and governments often resort to disparate measures, impeding the progress of the global energy and climate change agenda \cite{Fraune2018SustainableResistance}. Considering the urge to face this energy crisis, the integration of smart grids and the high penetration of renewable energy sources (RES) has been considered as the only way forward \cite{Su2021DoesRisks}.

The stochasticity of RES, along with the privatization and deregulation of electricity market pose severe uncertainties in electricity load forecasts. Additionally, energy demand patterns depend on numerous external variables such as the weather conditions, energy prices \cite{Spiliotis2021a}, seasonal factors and social activities leading to complex correlations and therefore demanding modelling for Load Forecasting (LF). The COVID-19 global health emergency is a bright example of an occasion that radically altered the patterns social activities for a long time and therefore power systems operation and electricity demand patterns \cite{Navon2021EffectsAhead}. These distribution shifts are projected on the patterns of time series and definitely require either dedicated handling depending on the nature of the utilized forecasting technique \cite{Obst2021AdaptiveFrance, Tudose2021Short-TermStudy}. This variety of factors and events, form a multiparametric and complex problem calling for innovative and rigorous LF solutions with out-of-distribution generalization capabilities \cite{Arjovsky2020OutLearning}. 

Fortunately, the smart grid paradigm is, essentially, wide open to innovative approaches enabling  the integration and exploitation of real-time and data-centric solutions powered by Information and Communication Technologies (ICT) such as Big Data, Internet of Things (IoT), Artificial Intelligence (AI) and lately Machine Learning Operations (MLOps) \cite{Alla2021} towards harmonized network operation \cite{Marinakis2020FromManagement, Marinakis2020BigBuildings}. Such technologies, and especially MLOps, have the potential to revolutionize the energy sector by establishing a new paradigm in load forecasting (LF) that goes beyond conventional model development processes. In this context, automated and continuous model training, evaluation, validation and deployment are expected to replace the manual processes of the conventional Machine Learning (ML) lifecycle, leading to faster and more efficient decision making for network operators and generally Electrical Power and Energy System (EPES) stakeholders. 

LF is divided into four major categories based on the forecasting horizon. Very short-term load forecasting (VSTLF) addresses forecast horizons of a few minutes to a few hours. It is mainly used for real-time control and demand response. Electricity demand forecasts of one day to one week ahead are known as Short-Term Load Forecasting (STLF). The resolution of such forecasts ranges from 15-minute to hourly and mainly serves for day-to-day operations of utilities and participation in the electricity market. Medium-term load forecasting (MTLF) involves demand forecasts of a week up to a year ahead and is mainly used for maintenance scheduling and planning of fuel purchases as well as energy trading and revenue assessment for the utilities. Finally, long-term load forecasting (LTLF) may involve time horizons up to 20 years ahead and is usually linked to grid development and strategical planning \cite{Hammad2020MethodsReview}. ML based STLF forms the core task of this study and therefore the focus of the rest of this paper is shifted towards this direction.

\subsection{Related Work}

Over the years, various forecasting approaches have been proposed in literature for load forecasting, and specifically Short-Term Load Forecasting (STLF). However, the studies have been inconclusive regarding the superiority of a specific technique as different solutions exhibit varying robustness, complexity and requirements for computational resources. Additionally, each time series and each forecasting technique exhibit their unique characteristics depending on the use case in question and the scope of the analysis. Forecasting models can be divided into two basic categories, namely: (a) conventional statistical methods and basic ML methods, b) advanced, state-of-the-art methods that usually include Deep Learning (DL) and, very commonly, hybrid setups. Conventional methods usually serve as baselines. 

Statistical models like linear regression have been widely used for STLF \cite{Moghaddas-Tafreshi2008a, Yildiz2017AForecasting}. Specifically, Dudek in \cite{Dudek2016Pattern-basedForecasting} combined them with pattern-based techniques. Simpler models like the Box-Jenkins method \cite{Vahakyla1980Short-termTechniques}, Exponential Smoothing (ES) \cite{Christiaansea}, and nonparametric regression \cite{Clements2016ForecastingApproach} are also common in literature. ARIMA models \cite{Li2009BasedForecasting} and their variations, such as ARIMAX \cite{Cui2015Short-TermModel} and SARIMA \cite{MusbahSARIMADetection}, have also been an option for STLF. Techniques like decomposition \cite{Mohan2018AModel} are rarely used alone, however they are often met in hybrid models \cite{Annamareddi2013AForecasting}, such as Wavelet transformation \cite{Du2002}. Conventional methods can also include simple ML approaches. These models usually are Feed-Forward Artificial Neural Networks (ANN) \cite{Fan2009Short-termInformation, Baczynski2004InfluenceForecast} and Support Vector Machines (SVM) \cite{Hong2011ElectricAlgorithm, Che2014Short-termModel}. Self-Organising Maps (SOM) \cite{Fan2006Short-termMethod}, regression trees \cite{Mori2001OptimalForecasting}, random forests (RF) \cite{Dudek2015Short-TermForests}, gradient boosting regression trees (GBRT) \cite{BenTaieb2014ACompetition} and Kalman filtering \cite{Al-Hamadi2006FuzzyFilter} have been also proposed for STLF. Ultimately, clustering approaches have proven to be very efficient in predicting load profiles \cite{Martinez-Alvarez2011EnergySimilarity}. Conventional models sometimes fail to model the nonlinear relationship between weather and load or are susceptible to outliers \cite{Deng2019Multi-scaleForecasting}, however they offer simplicity and robustness \cite{Song2005Short-termMethod}, and can serve as baselines.

More advanced and state-of-the art-models very often include hybrid and ensemble models. Genetic algorithms (GA) have been widely used in order to minimize training time \cite{Liao2006ApplicationForecasting, Chen2016Group-basedForecasting} and select training parameters for Genetic Algorithm Neural Network (GA-NN) hybrid models \cite{Vesa2020EnergyPrograms, Mishra2008ShortOptimization}. Fuzzy logic (FL) \cite{Song2005Short-termMethod, Al-Hamadi2006FuzzyFilter} ensembled with other models is used for better and faster handling of very long time-series \cite{Senjyu2005NextMethod}. Convolutional Neural Network (CNN)-Recurrent Neural Network (RNN) hybrids are often used to better correlate load patterns with weather conditions \cite{Sajjad2020AForecasting, Rafi2021ANetwork}. ANNs can be trained from an Artificial Immune System model mainly for hyper-parameter optimization \cite{Hernandez2013APlants, Farsi2021OnApproach}. Additional fields of research include extreme learning machine \cite{Zhang2013Short-termMachine}, sequence-to-sequence models and specifically encoder-decoder architectures \cite{Sehovac2020DeepAttention, Rueda2021Short-termGrid, Henselmeyer2021Short-termTraining}. These advanced techniques can also have various applications in more specific problems like transmission system loads for special days, such as New Years’ Eve or Easter \cite{Song2005Short-termMethod}, or even special loads like data center consumption \cite{Vesa2020EnergyPrograms}.

\subsection{Contribution}

The high-level objective of this paper is STLF at Transmission System Operator (TSO) level. Specifically, we examine day-ahead forecasts of 15-minute resolution (96 timesteps-ahead predictions) for the net load time series of the Portuguese transmission system. Specifically, the purpose of this paper is to perform an extensive comparison of DL architectures, namely Neural Basis Expansion Analysis Time Series Forecasting (N-BEATS) \cite{Oreshkin2019N-BEATS:Forecasting}, Long Short-Term Memory Recurrent NNs (LSTM) \cite{Hochreiter1997a} and Temporal Convolutional Networks  (TCN) \cite{Bai2018AnModeling}. An averaging and a stacking ensemble of PSF estimators \cite{Martinez-Alvarez2011EnergySimilarity} serve as baseline models. Moreover, we attempt to demonstrate the effects of the COVID-19 global emergency that initiated in early 2020 on the performance of the employed models by alternating their training, validation, and testing periods accordingly. Ultimately, motivated by the capabilities of the smart grid for integration of ICT technologies and automation we also propose a high-level model deployment scheme targeted for production environments and production systems of network operators.

\subsection{Structure of the Article}

The rest of the paper is organized as follows. Section \ref{sec:2} describes our methodological approach, including the utilized methods, architectures, and the proposed evaluation and deployment scheme. Section \ref{sec:3} summarizes the results of the forecasting models, in terms of accuracy and sustainability, also discussing the effects of the pandemic on them, followed by concluding remarks and future work in Section \ref{sec:4}.

\section{Methodology \label{sec:2}}
This section presents the followed methodology for addressing the common stages of an ML lifecycle. Specifically, the stages described are (i) data collection, wrangling and transformations; (ii) exploratory analysis of the dataset; (iii) selection of ML and DL architectures along with their development process and model validation; (iv) an evaluation framework tailored to the requirements of production smart grid environment and finally (v) a proposed conceptual framework for models’ deployment to production based on the principles of MLOps.

\subsection{Data Collection and Feature Extraction}
The main part of the dataset consisted of the Portuguese net energy demand time series at a 15-minute resolution. Such data are publicly available online. Except for performing a comparative analysis of state-of-the-art ML models in forecasting the Portuguese load at day-ahead level, the aim of this study is also to observe the effects of the COVID-19 pandemic on model performance. Therefore, collected net load data from 2009 to 2020 were utilized, as they permit to experiment with different whole year test sets that either referred (2020) or not (2019) to time periods related with the global health emergency. Some common operations that were applied before proceeding to transformations and feature engineering, are: (i) removal of duplicate entries (mainly caused by changes between standard and daylight saving time) and (ii) filling missing data (very few cases that were handled with linear interpolation).

Autoregressive models can predict future values of a time series by looking back at its past values. However, as already discussed, various exogenous variables can affect its patterns, therefore holding predicting power. Amongst them, we decided to model the time in an attempt to capture seasonal factors and special events. An initial step of data transformations was thus performed leading to additional variables, from now on referred to as temporal covariates. Temporal covariates are synthetic variables that encode the timely characteristics of each time step and can be used as predictors enriching the input feature space of the forecasting problem. Table \ref{tab:1} lists the specific temporal covariates accompanied by their modelling approach. A very common encoding approach is through sinusoidal and cosinusoidal transformations as they can reproduce the continuousness periodicity of time. Both transformations need to be included in the final temporal covariates to avoid getting the same values for different time moments due to their nonmonotonic nature.

\begin{table}[htbp]
\caption{Characteristics of Introduced Temporal Covariates}
\begin{center}
\begin{tabular}{|c|c|c|c|}
\hline
\textbf{Time}&\textbf{Periodicity}&\textbf{Encoding}&\textbf{Number of } \\
\textbf{Feature}& & \textbf{Method} & \textbf{Derived Covariates} \\
\hline
        Year & - & Integer & 1  \\ \hline
        Month  & Yearly & Sine \& Cosine & 2  \\ \hline
        Day of Year & Yearly & Sine \& Cosine & 2  \\ \hline
        Day of Week & Weekly & Sine \& Cosine & 2  \\ \hline
        Week of Year & Yearly & Sine \& Cosine & 2  \\ \hline
        Holiday & Yearly & Boolean & 1  \\ 
\hline

\end{tabular}
\label{tab:1}
\end{center}
\end{table}

\subsection{Exploratory Data Analysis}
\begin{figure}[htbp]
\centerline{\includegraphics[height=5.5cm, keepaspectratio]{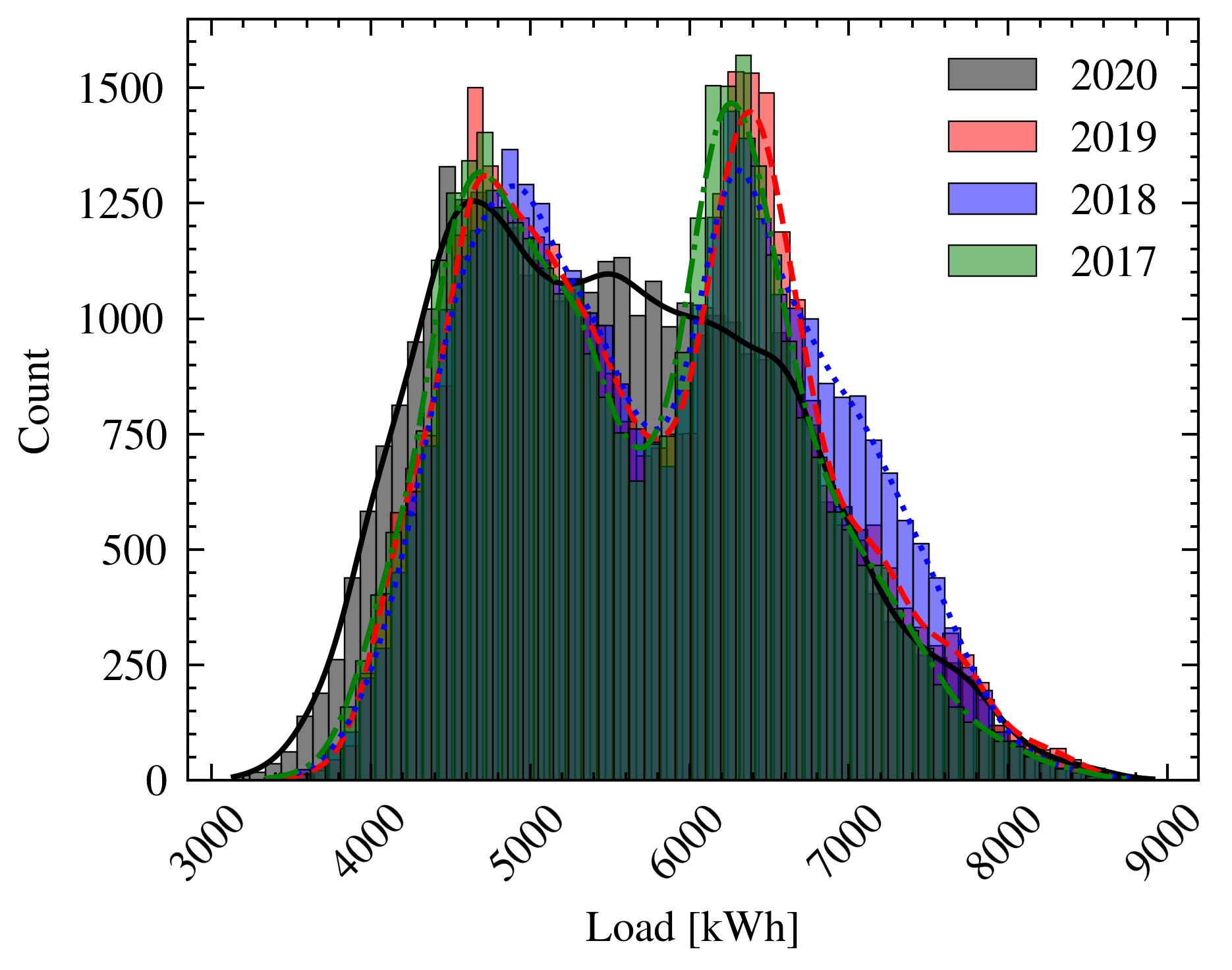}}
\caption{Histogram of the 15-minute load values for the years 2017-2020.}
\label{fig:1}
\end{figure}

\begin{figure}[htbp]
\centerline{\includegraphics[width=\linewidth, height=5.5cm, keepaspectratio]{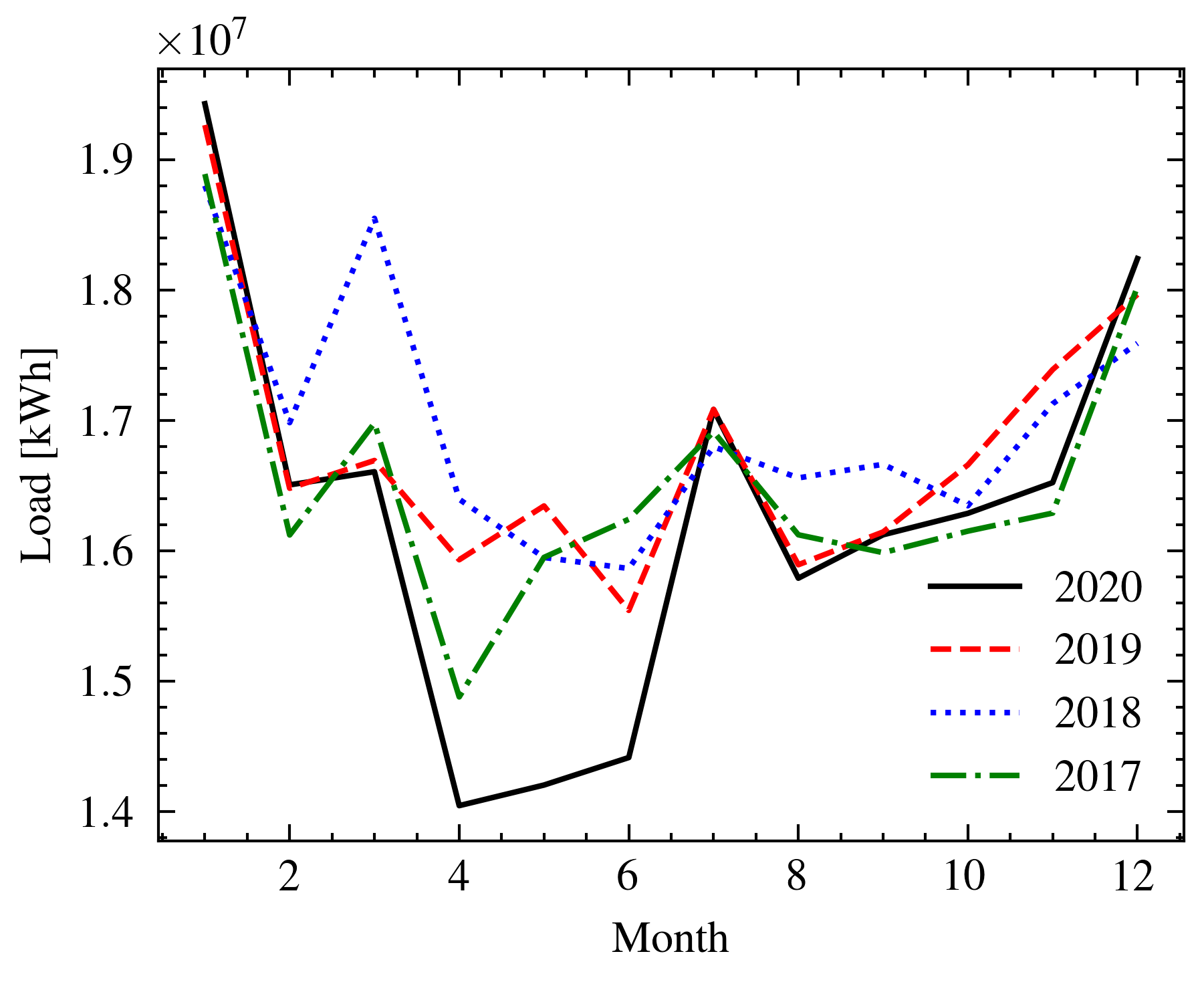}}
\caption{Average monthly load for the years 2017-2020.}
\label{fig:2}
\end{figure}

\begin{figure}[htbp]
\centerline{\includegraphics[width=\linewidth, height=5.5cm, keepaspectratio]{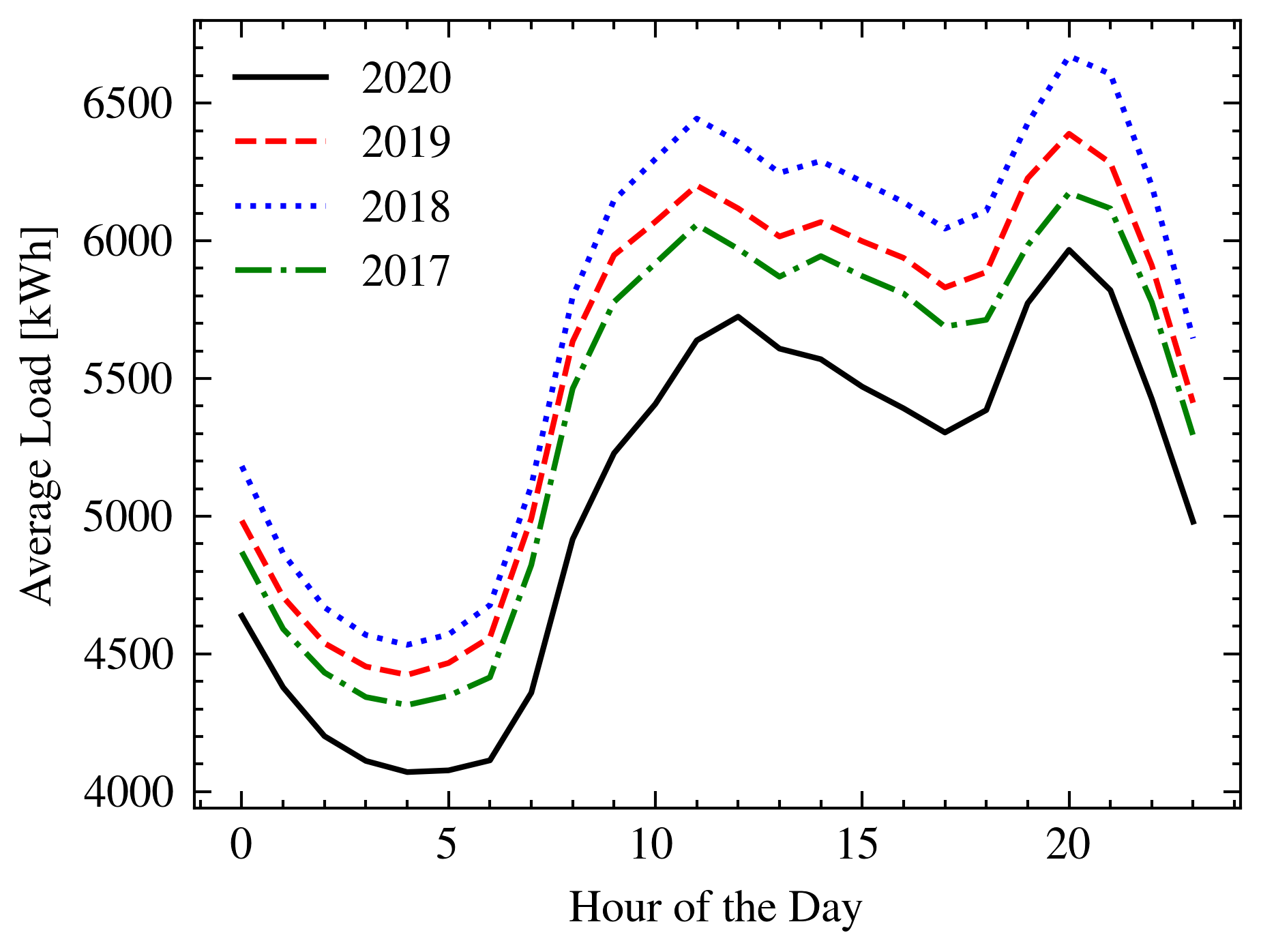}}
\caption{Average daily load profiles for the years 2017-2020.}
\label{fig:3}
\end{figure}

Following data wrangling and feature engineering, several data exploratory analytics and insights were extracted for the dataset. These analytics mainly contribute to visualize and understand the basic patterns of the load series along with the pandemic’s effects on the time series profile, motivating the COVID-19 related analysis. All statistics range from 2017 to 2020, highlighting the distribution shift of the load time series during the lockdown period, which began on $17^{th}$ of March 2020 and lasted until the 3rd of May 2020 when a gradual relaxation of measures was applied. Fig. \ref{fig:1} illustrates the alteration of the distribution of load values within 2020 compared to previous years, denoting a general load reduction, smoothing and transition to a unimodal rather than bimodal yearly distribution. Fig. \ref{fig:2} illustrates the total monthly load values for all years of interest, pinpointing a harsh and unprecedented decrease during the lockdown which is highly expected due to the commercial, tourism and food industries of the country being closed and the corporate ones shifting to the Work from Home paradigm. Fig. \ref{fig:3} clearly depicts the alteration of average daily profiles with respect to mostly their magnitude yet their shape as well. In summary, all observations imply that any conventional model being trained on the historical values of the Portuguese load series is expected to fail to completely adapt to the distributions shift caused by the pandemic, leading to dips in performance and increased errors while predicting 2020. This factor subjects to further investigation within the present study.

\subsection{Model Selection}
In this section we present the ML and DL models employed within our experimentation.

\textbf{PSF.} The Pattern Sequence Forecast algorithm (PSF) introduced by Martinez-Alvarez et al. in \cite{Martinez-Alvarez2011EnergySimilarity}, is a univariate time-series forecasting approach based on the similarity between pattern sequences. It consists in two distinct steps: the clustering step, in which time-series patterns are grouped, and the predictive step, which leverages on the cluster models to generate forecasts. PSF therefore treats training data as a set of cluster labels, using their sequencing to find and match windows of similar sequencing and combining them into a single forecast. 

\textbf{LSTM.} Unlike standard feedforward neural networks, RNNs have feedback connections that render them appropriate for processing sequential data. Long short-term memory (LSTM) is an extension of the conventional RNN that came to solve the vanishing gradient problem, mitigating the inability to learn long range dependencies \cite{Hochreiter1997a}. LSTMs have been widely used for short-term load forecasting tasks. A very common approach of employing LSTMs is through encoder-decoder setups—instead of fully recurrent architectures—especially for sequence-to-sequence tasks, and therefore were utilized for the multi-step horizon forecasts of the present study (96 forecast steps) \cite{Sehovac2020DeepAttention, Rueda2021Short-termGrid, Henselmeyer2021Short-termTraining}.

\textbf{N-BEATS} is a neural network that was first described in a 2019 article by Oreshkin et al. in \cite{Oreshkin2019N-BEATS:Forecasting}. The authors reported that N-BEATS outperformed the M4 forecast competition winner by 3\%. The M4 winner was a hybrid between a recurrent neural network and Holt-Winters exponential smoothing, whereas N-BEATS implements a “pure” deep neural architecture. N-BEATS treats forecasting as a non-linear multivariate regression problem. It comprises a very deep stack of fully connected non-linear neural regressor blocks interconnected with doubly residual links. N-BEATS has already been validated on mid-term energy forecasting tasks \cite{Oreshkin2021a}. However, our case study tasks fall within the scope of STLF, hence opening a new perspective for validating the architecture on this task.

\textbf{TCN} is a CNN that  consists of dilated, causal 1-dimensional convolutional layers with the same input and output lengths and was first proposed in \cite{Bai2018AnModeling} as a solution for time series forecasting tasks. The authors reported that a simple convolutional architecture is more effective across diverse sequence modelling tasks than recurrent architectures such as LSTMs and GRUs. TCNs have been recently validated for STLF tasks \cite{Yin2021Multi-temporal-spatial-scaleSystems, Tang2022Short-TermNetwork} and are employed throughout our case study. 

\subsection{Model Training}
The aim of the training process has been the creation of models for day-ahead forecasting at 15-minute resolution for the net electricity demand of the Portuguese transmission system. In this context, the selected ML and DL methods were trained for multiple combinations of hyperparameters allowing to select the optimal architecture for each method, as described in detail in Table \ref{tab:2}. Additionally to the hyperparameters of the table, an experimentation on different lookback window values took place, namely 384 (3 days), 672 (7 days), and 960 (10 days) timesteps. Note that the number of layers of the TCN architectures are set to “auto” as they are adjusted at each training round to ensure full receptive field coverage based on Eq. \ref{eq:1}\footnote{https://unit8.com/resources/temporal-convolutional-networks-and-forecasting/}.
\begin{equation}
\label{eq:1}
n=\log _{b}\left(\frac{(l-1) \cdot(b-1)}{(k-1)}+1\right)
\end{equation}
where $n$ represents the number of convolutional layers, $l$ the length of the lookback window, $b$ the dilation base, and $k$ the kernel size. For the training of the PSF baseline models several instances were obtained for different pattern window lengths and underlying clustering algorithms, including K-Means, SOM, and Gaussian Mixture Models. Each of the baseline algorithms was optimized using the Silhouette Index. The final optimized PSF models were used to create ensemble forecasts, using both averaging and stacking ensemble techniques, with SVR as the meta-learning algorithm. Model training took place on a laptop with an AMD Ryzen 9 4900H CPU, an NVIDIA RTX2060 6GB GPU, and 24GB RAM.

\begin{table}[htbp]
\caption{Model Flavors Ordered By Increasing Complexity Of Deep Learning Architecture Hyperparameters}
\begin{center}
\begin{tabular}{|c|c|c|c|}
\hline
\textbf{Base}&\multicolumn{3}{|c|}{\textbf{Architectural Properties per Model Flavor}} \\
\cline{2-4} 
\textbf{Model} & \textbf{\textit{Property Name}}& \textbf{\textit{Flavor 0}}& \textbf{\textit{Flavor 1}} \\
\hline
        N-BEATS & Number of stacks & 20 & 30  \\ \cline{2-4} 
        ~ & Number of blocks & 1 & 1  \\ \cline{2-4} 
        ~ & Number of layers & 4 & 4  \\ \cline{2-4} 
        ~ & Layer widths & 64 & 512  \\ \cline{2-4} 
        ~ & Expansion coefficient dim. & 5 & 5  \\ \hline
        Encoder –  & Recurrent layers  & 1 & 2  \\ \cline{2-4} 
        Decoder LSTM & Hidden dimension & 20 & 64  \\ \cline{2-4} 
        ~ & Dropout & 0 & 0  \\ \cline{2-4} 
        ~ & Learning rate & 0.0008 & 0.001  \\ \hline
        TCN & Kernel size & 3 & 5  \\ \cline{2-4} 
        ~ & Number of filters & 3 & 5  \\ \cline{2-4} 
        ~ & Dilation base & 2 & 3  \\ \cline{2-4} 
        ~ & Convolutional layers & auto & auto  \\ \hline
        PSF  & Ensemble Method & Averaging & SVR  \\
\hline
\end{tabular}
\label{tab:2}
\end{center}
\end{table}

The dataset split was as follows: (i) training set: 2009-2017 (ii) validation set: 2018 (iii) test set: 2019. Subsequently, the best models from the previous step were retrained to investigate the effect of the COVID-19 load distribution shifts on model performance. For this purpose, the dataset split was modified for the test set to include the pandemic period as follows: (i) training set: 2009-2018 (ii) validation set: 2019 (iii) test set: 2020. The results of the experiments are summarized in the section \ref{sec:3}.

With respect to temporal covariates, the N-BEATS model was trained without them as: (i) indicated in \cite{Oreshkin2019N-BEATS:Forecasting, Oreshkin2021a}; (ii) it was observed that its performance was systematically lower when the covariates were included in the model. On the contrary, LSTMs and TCNs are trained and evaluated including the temporal covariates for the opposite reason. Temporal covariates are not applicable to the PSF method. 

\subsection{Model Validation and Evaluation Framework}
Regarding model validation and evaluation, in our case every DL model was: (i) trained on the predefined training set; (ii) validated on the validation set to discover the best hyperparameters; (iii) and evaluated on the remaining and previously unseen test set. With respect to performance measures, various forecast errors are common in forecasting applications including the Mean Absolute Error (MAE), the Mean Squared Error (MSE), the Mean Absolute Percentage Error (MAPE), the Root Mean Square Error (RMSE), the Symmetric Mean Absolute Percentage Error (SMAPE), and the Mean Absolute Scaled Error (MASE) \cite{Hyndman2006}. We use MAPE (Eq. \ref{eq:2}) as a widely accepted choice for STLF benchmarks.
\begin{equation}
MAPE=\frac{1}{m} \sum_{i=1}^{m}\left|\frac{Y_{t}-F_{t}}{Y_{t}}\right| \cdot 100(\%)
\label{eq:2}
\end{equation}
where $m$ represents the number of samples, $Y_t$ and $F_t$ stand for the actual data values and forecasted data values, respectively. 

Regarding the calculation of MAPE—as our experimentation is aimed for smart grid applications and deployment to production—we opted for a backtesting approach. Backtesting refers to the periodical updating of a model’s history depending on its forecasting horizon. To elaborate more, let us suppose a TSO that aims to participate in the Portuguese daily energy market. Every day, a 96-step-ahead (day-ahead) forecast is required. However, at the end of the forecasted day, the ground truth values of the series are already known and thus can be fed to the model to update its lookback window for day-ahead inference. Hence, the proposed DL models are validated and evaluated following this approach, which is however not applicable to the PSF method.

\subsection{High-Level Deployment Framework}
\begin{figure}[htbp]
\centerline{\includegraphics[width=\linewidth, height=7.5cm]{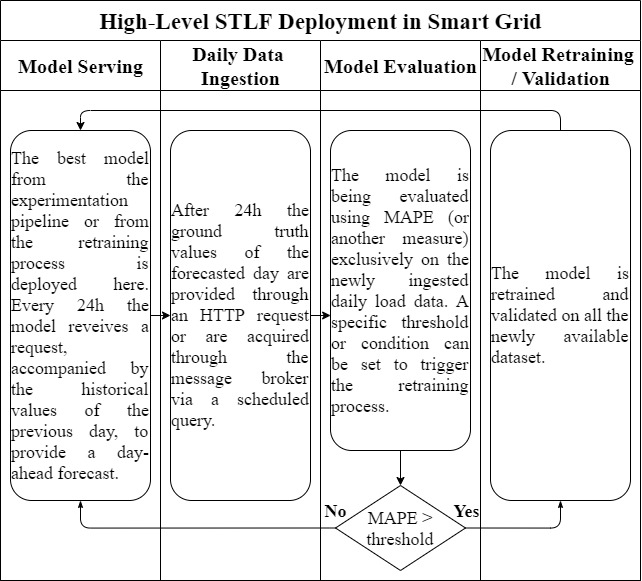}}
\caption{High-level deployment framework for the STLF pipeline.}
\label{fig:4}
\end{figure}

As previously discussed, the proposed deployment framework can be part of a TSO smart grid system that is used for producing daily load forecasts in an automated manner. At this deployment stage, it is important that the model’s performance is constantly monitored by being evaluated on newly ingested data. Dips in performance possibly indicate that the entire training process may need to be repeated to update the model to incorporate new trends. Fig. \ref{fig:4} illustrates this concept within a smart grid application.

\section{Results \label{sec:3}}
This section presents the results of our study. Python and MLflow \cite{Alla2021} were used to set up the experimentation pipeline and keep track of of the performed experiments.

\subsection{Results of the Comparative Analysis}
The results of the comparative analysis for the 2009-2019 experimentation period (training set: 2009-2017, validation set: 2018, test set: 2019) are shown in Table \ref{tab:3}. They refer both to model accuracies (MAPE) and GPU training times as such a measurement can be a good indicator of environmental friendliness and sustainability. Specifically, the N-BEATS architecture clearly outperforms the LSTM and TCN architectures from both perspectives with Model 4 achieving the lowest MAPE at a relatively low GPU time. Moreover, regarding TCN and LSTM it can be concluded that a lookback window (l) of 7 days (672 15-minute timesteps) is the best option for higher accuracy. However, this is not the case for N-BEATS which performs better for the 10-day lookback window. Note here that all deep architectures outperform the PSF baseline model in terms of accuracy at the cost—of course—of significant GPU utilization.

As a general comment, larger and more complex neural network architectures (flavor 1) on average result to fewer epochs as they can fit the datasets faster, denoting lower GPU times and thus higher sustainability. In fact, except for N-BEATS, deeper architectures lead as well to higher accuracies, avoiding potential overfitting thanks to the early stopping approach. Note here that a poor selection of neural architecture—such as the flavor 0 of the LSTM family—can lead to low accuracy and increased GPU times and eventually higher electricity consumption during training. 

\begin{table}[htbp]
\caption{Model Training And Evaluation Results for the 2009-2019 period}
\begin{center}
\begin{tabular}{|c|c|c|c|c|c|c|}
\hline
~ & ~ & \multicolumn{2}{|c|}{\textbf{Architecture}} & \multicolumn{3}{|c|}{\textbf{Performance}} \\
\cline{3-7} 
\textbf{Model} & \textbf{Model} & ~ & ~ & ~ & ~ & \textbf{\textit{GPU}} \\
\textbf{Family} & \textbf{ID} & \textbf{\textit{Flavor}} & \textit{l} & \textbf{\textit{MAPE}} & \textbf{\textit{Epochs}} & \textbf{\textit{Time}} \\
~ & ~ & ~ & ~ & ~ & ~ & \textbf{(\textit{h})} \\
\hline
PSF & 0 & 0 & - & \textbf{3.333}   & - & \textbf{0}  \\ \cline{2-7}
~ & 1 & 1 & - & 3.581 & - & \textbf{0}  \\ \hline
N-BEATS & 2 & 0 & 384 & 2.245 & 51 & 1.10  \\ \cline{2-7}
~ & 3 & 0 & 672 & 2.212 & 43 & 1.00  \\ \cline{2-7}
~ & 4 & 0 & 960 & \textbf{2.001} & 71 & 1.49  \\ \cline{2-7}
~ & 5 & 1 & 384 & 2.305 & 23 & \textbf{0.94}  \\ \cline{2-7}
~ & 6 & 1 & 672 & 2.315 & 31 & 1.04  \\ \cline{2-7}
~ & 7 & 1 & 960 & 2.179 & 37 & 1.18  \\ \hline
LSTM & 8 & 0 & 384 & 2.437 & 261 & 5.85  \\ \cline{2-7}
~ & 9 & 0 & 672 & 2.57 & 141 & 3.67  \\ \cline{2-7}
~ & 10 & 0 & 960 & 2.56 & 149 & 3.85  \\ \cline{2-7}
~ & 11 & 1 & 384 & 2.212 & 35 & \textbf{1.19}  \\ \cline{2-7}
~ & 12 & 1 & 672 & \textbf{2.087} & 43 & 1.98  \\ \cline{2-7}
~ & 13 & 1 & 960 & 2.201 & 41 & 2.15  \\ \hline
TCN & 14 & 0 & 384 & 2.525 & 127 & 3.06  \\ \cline{2-7}
~ & 15 & 0 & 672 & 2.409 & 145 & 3.44  \\ \cline{2-7}
~ & 16 & 0 & 960 & 2.413 & 117 & 3.04  \\ \cline{2-7}
~ & 17 & 1 & 384 & 2.293 & 123 & 2.70  \\ \cline{2-7}
~ & 18 & 1 & 672 & \textbf{2.243} & 99 & 2.58  \\ \cline{2-7}
~ & 19 & 1 & 960 & 2.268 & 89 & \textbf{2.39}  \\ \hline
\end{tabular}
\label{tab:3}
\end{center}
\end{table}

\subsection{Impact of the Pandemic on the Model Performance}
\begin{figure*}[htbp]
\centerline{\includegraphics[width=\linewidth, height=7cm, keepaspectratio]{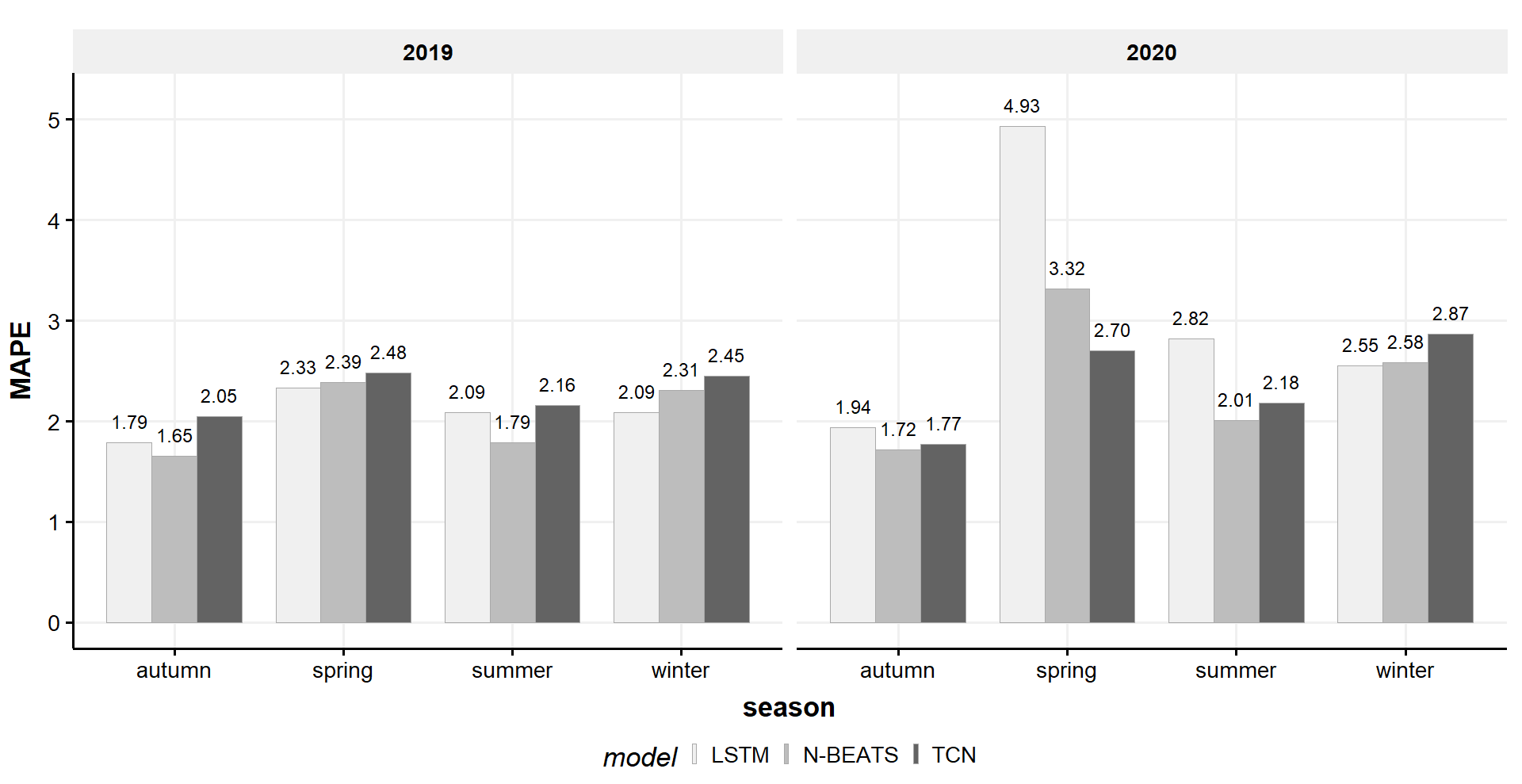}}
\caption{Model errors per season for the years 2019 (normal year) and 2020 (COVID-19 global health emergency)}
\label{fig:5}
\end{figure*}

To examine the impact of the COVID-19 global health emergency on DL models' performance, the most accurate setups from each DL model family from the experiments of 2009-2019 period (Model 4, 12 and 18) were retrained on 2009-2018 validated on 2019, and tested on 2020. Even though the training dataset is larger by one year, a condition that is usually beneficial for the accuracy of DL models, this is not the case here as most models perform lower on average as illustrated in Fig. \ref{fig:5}. To get a more detailed insight on the effects of the pandemic, the model errors are listed for each season separately, revealing the most affected period by the pandemic. Expectedly, during the spring and summer periods the MAPE is significantly increased. Of course, the lockdown period, which can be approximately projected to spring 2020 (also accounting for the post-lockdown measures) is linked with larger errors, especially for the LSTM and N-BEATS models. Nonetheless, the TCN model exhibits a much smaller accuracy decrease proving to be the most robust choice for the distribution shift observed during the lockdown period. A potential interpretation of this result can be that the TCN model exhibited a relative "underfitting" compared to the other models for the testing period of 2019; and underfitted models usually tend to estimate less accurately the target distribution, however they are less affected by potential distribution shifts, which is the case for 2020.

\section{Conclusions and Future Work \label{sec:4}}
Power grids have been undergoing an unprecedented change during the last few years, integrating RES and new technologies such Big Data and ML. In this context, STLF is a multiparametric task of great importance as it allows for harmonized grid operation and participation of EPES stakeholders in the energy market. In the meanwhile of this smart grid evolution, the COVID-19 global health emergency triggered socioeconomic phenomena that led to distribution shifts of energy demand, demonstrating the need for rigorous forecasting models and out-of-distribution generalization. The present work focuses on comparing three state-of-the-art DL architectures—namely N-BEATS, encoder-decoder LSTM, and TCN—in terms of accuracy and sustainability, also examining their robustness during the pandemic period. 

During the pre-pandemic period, it has been demonstrated that N-BEATS outperforms both the other architectures leveraging a relatively simple architecture, while also exhibiting acceptable GPU times and hence sustainability. Nonetheless, it is significantly affected by the load distribution shifts of the pandemic leading to decreased accuracy when tested on 2020. The TCN architecture exhibits lower accuracy and sustainability in general, however it demonstrates better out-of-distribution generalization during the pandemic, as a potential result of underfitting the pre-pandemic training set. The encoder-decoder LSTM proved to be highly sensitive to the architecture depth and complexity, while it also failed to adapt to the COVID-19 distribution shift. As a main conclusion, it is always upon EPES stakeholders and their ML engineers to decide on their architecture of preference for STLF tasks based on their computational resources, error tolerance per season, and readiness for model retraining in case of distribution shifts. Hence, we would suggest TCN for those that need a "one-for-all" solution without the constant necessity for monitoring and retraining. On the contrary, we would suggest N-BEATS for stakeholders that require state-of-the-art performance and high sustainability at the cost of an automated monitoring process as the one proposed in the deployment scheme of Fig. \ref{fig:4}.\newline \indent Future work includes: (i) the integration of weather variables—and specifically weather forecasts—in the models; (ii) an exhaustive hyperparameter space search as long as High-Performance Computers (HPC) become available; (iv) the extensive search of rigorous setups—including DL ensembles—that will allow for out-of-distribution generalization without the compromise of underfitting during normal periods; (v) a thorough study on the tradeoff between accuracy and GPU times in terms of grid and environmental sustainability. From a broader perspective, a future step could as well be the extension of this experimentation to more European countries alongside the development of a production ICT system for automated and easily deployable STLF.

\section*{Acknowledgment}

This work has been funded by the European Union’s Horizon 2020 research and innovation programme under the I-NERGY project, Grant Agreement No 101016508. 

\printbibliography

\end{document}